\documentclass[a4paper, 10pt, conference]{ieeeconf}

\newif\ifarwfinalcopy

\arwfinalcopytrue  

\IEEEoverridecommandlockouts                       

\usepackage{graphics} 
\usepackage{epsfig} 
\usepackage{mathptmx} 
\usepackage{times} 
\usepackage{amsmath} 
\usepackage{amssymb}  
\usepackage{subcaption}
\usepackage{lineno}
\usepackage[T1]{fontenc}
\usepackage{url}
\usepackage{tikzpagenodes}
\usepackage{background}
\usepackage{multirow}

\ifarwfinalcopy
\backgroundsetup{color=white}
\else

\linenumbers

\SetBgContents{\MyARWConfidentialLogo}
\SetBgPosition{current page.north west}
\SetBgOpacity{0.5}
\SetBgAngle{0.0}
\SetBgScale{1.0}

\fi

\title{\LARGE \bf
Sim2Real Transfer for Vision-Based Grasp Verification
}

\author{Pau Amargant$^{1,2}$, Peter Hönig$^{1}$, and Markus Vincze$^{1}$
\thanks{$^{1}$ All authors are with Faculty of Electrical Engineering, Technical University of Vienna, 1040 Vienna, Austria; {\tt\small \{hoenig, vincze\}@acin.ac.tuwien.at}}%
\thanks{$^{2}$ Pau Amargant is with Polytechnic University of Catalonia; {\tt\small pau.amargant@estudiantat.upc.edu}}
}

\begin{document}

\maketitle

\begin{abstract}
The verification of successful grasps is a crucial aspect of robot manipulation, particularly when handling deformable objects. Traditional methods relying on force and tactile sensors often struggle with deformable and non-rigid objects. 
In this work, we present a vision-based approach for grasp verification to determine whether the robotic gripper has successfully grasped an object.
Our method employs a two-stage architecture; first a YOLO-based object detection model to detect and locate the robot's gripper and then a ResNet-based classifier determines the presence of an object.
To address the limitations of real-world data capture, we introduce HSR-GraspSynth, a synthetic dataset designed to simulate diverse grasping scenarios.
Furthermore, we explore the use of Visual Question Answering capabilities as a zero-shot baseline to which we compare our model.
Experimental results demonstrate that our approach achieves high accuracy in real-world environments, with potential for integration into grasping pipelines. Code and datasets are publicly available at \url{github.com/pauamargant/HSR-GraspSynth}
\end{abstract}


\section{INTRODUCTION}
Deformable object manipulation is a growing field of research in robotics due to its relevance in a wide range of tasks~\cite{zhu_challenges_2022}.
Deformable objects are a common occurrence in both industrial and household environments, and their manipulation poses challenges when compared to rigid objects. 
Their deformation and varying response to traditional force and tactile sensing methods during the grasping process introduce significant uncertainty, making it a more challenging task~\cite{yin_modeling_2021}. 

One critical aspect of deformable object manipulation is the verification of successful grasping. Traditional methods~\cite{bicchi_robotic_2000}, which often rely on the object's geometry and force and tactile sensors, struggle to account for the deformation of the object and its lack of internal structure and resistance~\cite{sanchez_robotic_2018}. This requires the use of more advanced sensors and control algorithms, which are often robot and situation specific.

In this context, computer vision has emerged as a promising tool to address these challenges. Various methods have been proposed to use 2D and 3D vision during the grasping process for tasks such as rope and cloth manipulation~\cite{nair_combining_2017, sun_accurate_2015}. 
These approaches use vision in combination with other input modalities such as tactile sensing to estimate the object's deformation during the grasping procedure. 
However, most proposed methods focus on the grasping control feedback and are object and task specific. 
These constraints and their complexity make these models unsuitable for the task of verifying a successful grasp.

This paper explores the application of computer vision for verifying whether a robot gripper has successfully grasped an object, with a focus on methods applicable to deformable objects. Our approach, which can be easily adapted to different robots and tasks, leverages object detection and machine learning to detect the grasping using the robot's on device camera. Our main contributions are as follows: 
\begin{enumerate}
\item We introduce a two-stage vision-based grasp verification model combining YOLO-based object detection and ResNet-based classification, improving generalization across different robotic platforms and object types.
\item We present HSR-GraspSynth, a synthetic dataset designed to simulate diverse grasping scenarios, addressing the limitations of real-world data collection and annotation.
\item We investigate the integration of Multimodal Large Language Models (LLMs) with Visual Question Answering (VQA) capabilities as a viable alternative for zero-shot learning in grasp verification.
\end{enumerate}



\section{RELATED WORK}
Deformable object manipulation is an active area of research in robotics with wide practical applications~\cite{zhu_challenges_2022}. 
Non-rigid objects are common in both industrial and domestic settings, making robots that can handle them especially useful.
However, manipulating them poses additional challenges compared to rigid objects.

One of the most important aspects of the grasping pipeline is the ability to verify its success.
Traditional methods use object geometry, force, tactile sensors~\cite{romano_human-inspired_2011}, and proximity sensors~\cite{kulkarni_low-cost_2019}, but often struggle to account for possible deformations and the lack of internal structure in deformable objects.

In this context, computer vision has been successfully applied to these challenges. 2D and 3D vision methods have been proposed for tasks like rope~\cite{nair_combining_2017} and clothing manipulation~\cite{sun_accurate_2015}. 
These methods combine vision with other modalities to determine grasping poses and account for the object's deformation during the procedure.  

However, the majority of these methods are focused on grasping estimation and the control feedback and are robot, object and task specific.  
Computer vision solutions have been proposed as a simpler alternative for the task of verification~\cite{nair_performance_2020}.
In 2020, the use of low-cost machine vision cameras installed in the robot gripper was studied~\cite{nair_performance_2020}.
They trained both YOLO~\cite{redmon_you_2016} and MobileNet~\cite{howard_mobilenets_2017} models for this task, achieving high precision with different camera systems.

Inspired by this approach, we aim to develop a similar solution for robots with head-mounted cameras, such as the PAL Robotics Tiago, Toyota HSR, and Boston Dynamics' Atlas, enabling vision-based grasp verification without relying on gripper-mounted cameras.


\subsection{Synthetic Datasets}
Synthetic datasets have become increasingly popular for object grasping research~\cite{newbury_deep_2023}. 
With the advent of deep neural networks, the large amounts of data required make the use of synthetic data an attractive alternative to the laborious task of acquiring and annotating real world data.

Synthetic datasets have been widely used for training computer vision models for object detection and robotic tasks. Tools such as GraspIt~\cite{miller_graspit_2004} and BlenderProc~\cite{denninger_blenderproc_2019} can be used to generate large-scale, photorealistic, and physics-aware datasets. 
By using commonly used object datasets such as YCB-V~\cite{xiang_posecnn_2018} and  \textit{ShapeNetV2}~\cite{chang_shapenet_2015}, synthetic data makes it possible to efficiently train models in zero-shot situations or when real data is costly to obtain. 
Common use cases are object detection and pose estimation \cite{lin_explore_2023, tremblay_falling_2018}.  

However, while there is a wide availability of synthetic datasets for grasp planning, there is a lack of datasets specifically designated for grasp verification.  Current datasets do not capture the nuances of successful and failed grasps, such as occlusions, edge cases, and variations in sensor perspectives. This highlights a significant gap in the field and the need for dedicated synthetic datasets to support research in grasp verification.

\section{HSR-GraspSynth dataset}
Training a robust and generalizable model for grasp verification requires a diverse and extensive dataset. 
However, collecting real-world data is often expensive and time-consuming, making synthetic data an attractive alternative. 
Synthetic data should be diverse and similar enough to the real-world distribution in order to minimize the Sim2Real gap.~\cite{tobin_domain_2017, weibel_measuring_2021}. 

With these goals in mind, we created the HSR-GraspSynth dataset for grasp verification. It consists of annotated RGB images, referred to as  \textit{examples}, showing the HSR robot's gripper from the perspective of its head-mounted camera.
Each example is annotated with a bounding box around the visible  parts of the gripper and a binary label indicating whether an object is present in it (\textit{object} or \textit{no\_object}).

Synthetic examples are generated from 3D simulated \textit{scenes}, where a full environment including the robot and background distractors is randomly configured. 
Several examples are generated from the same scene, forming a \textit{batch}.

The dataset consists of 12.000 examples and a separate validation dataset composed of 5.000 images.

\subsection{Data Generation}

Synthetic data is generated using BlenderProc, a procedural pipeline that integrates Blender within Python to facilitate the rendering of large datasets.

For each batch of the dataset, a new scene is generated. A model of the robot is positioned at the centre of an enclosed room, with its arm extended in front of its head. 
To simulate realistic environments and obtain robust models, between 2 and 15 distractor objects are randomly scattered within the field of view of the robot using a physics-based algorithm to ensure physically plausible poses and varied object scales. 

For training examples, distractors are sampled from the ShapeNetV2 dataset, while for the validation set, objects from the YCB-V dataset are used.

Ten examples are generated per batch to improve computational efficiency and mitigate some of BlenderProc's limitations.

For each example within a batch, the robot's arm's pose is randomized by perturbing the positions of the arm joints and the camera orientation. 
A randomly sampled object is then placed within the robot's gripper with probability 0.5 to generate both \textit{object} and \textit{no\_object} examples.
The grasped object is sampled from the same dataset used for the distractor objects.

When an object is placed between the gripper fingers, the object is first moved away from the robot to avoid collisions, and the gripper fingers are partially closed to make contact with the object. 
A convex hull approximation is used to detect when the object collides with the gripper fingers.  
The gripper fingers are slowly closed until a collision is detected. 

\begin{figure}[t]
   \centering
    \begin{subfigure}{0.15\textwidth}
        \centering
        \includegraphics[width=\linewidth]{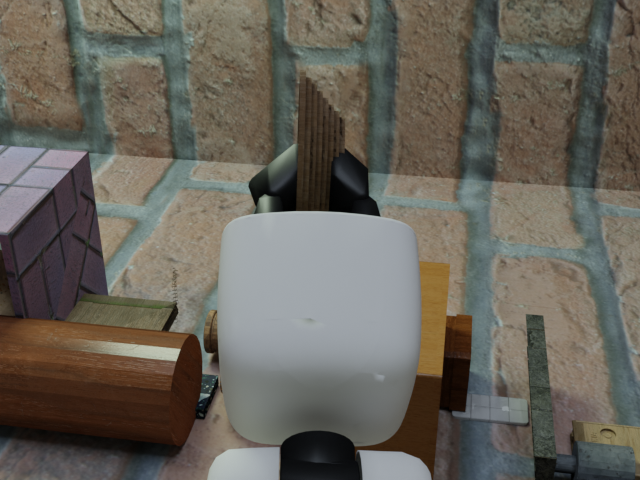}
    \end{subfigure}
    \hfill
    \begin{subfigure}{0.15\textwidth}
        \centering
        \includegraphics[width=\linewidth]{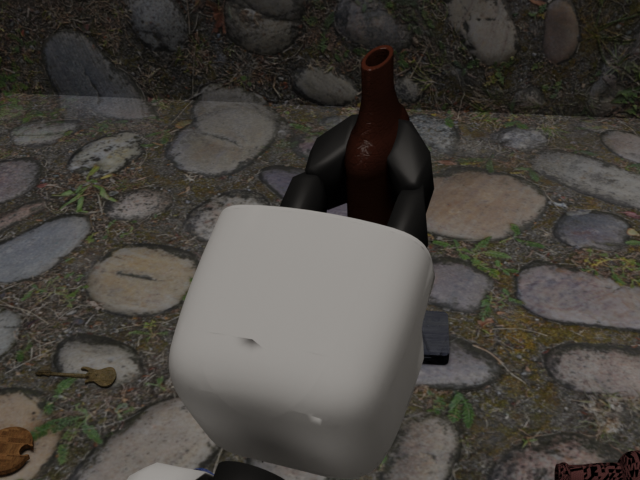}
    \end{subfigure}
    \hfill
    \begin{subfigure}{0.15\textwidth}
        \centering
        \includegraphics[width=\linewidth]{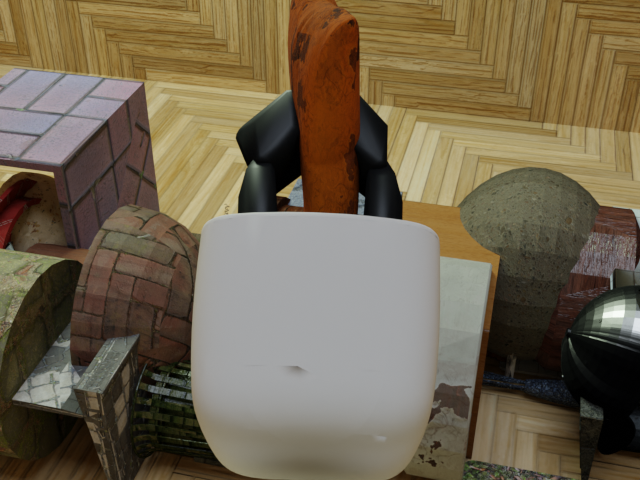}
    \end{subfigure}

    \vspace{0.5cm} 

    \begin{subfigure}{0.15\textwidth}
        \centering
        \includegraphics[width=\linewidth]{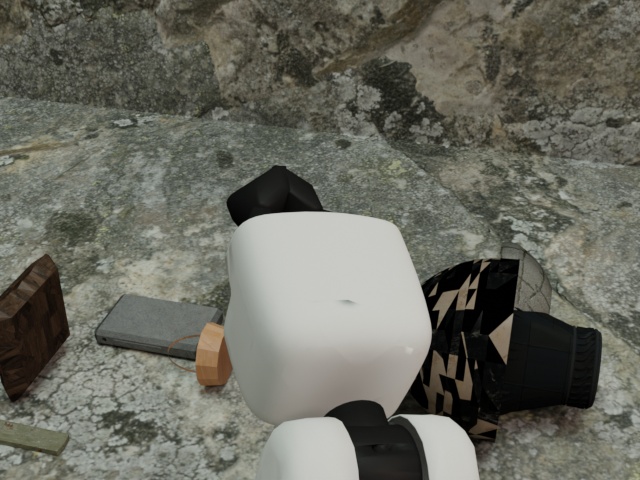}
    \end{subfigure}
    \hfill
    \begin{subfigure}{0.15\textwidth}
        \centering
        \includegraphics[width=\linewidth]{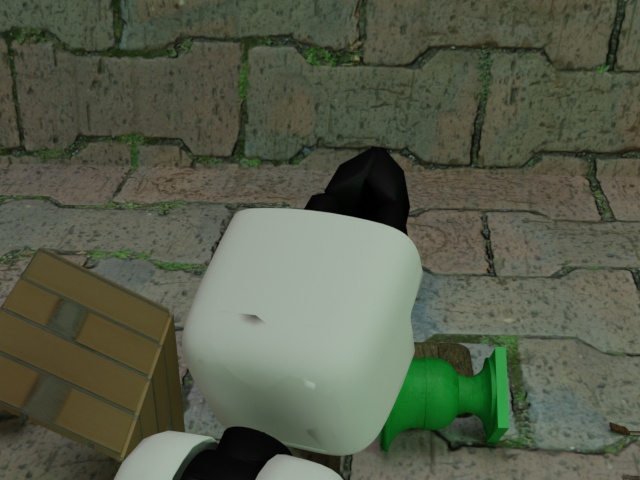}
    \end{subfigure}
    \hfill
    \begin{subfigure}{0.15\textwidth}
        \centering
        \includegraphics[width=\linewidth]{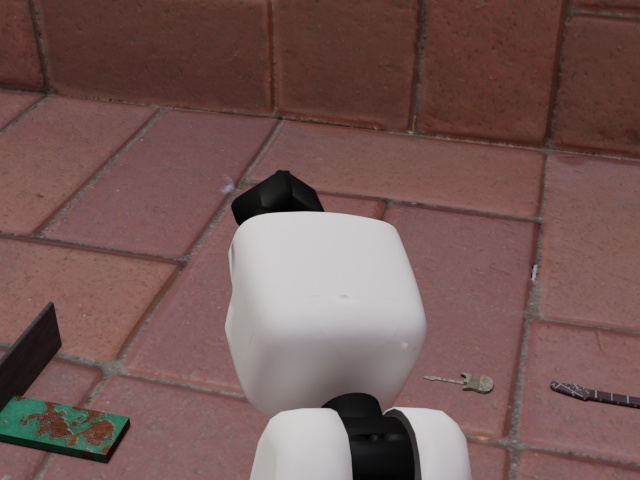}
    \end{subfigure}
    \caption{Examples from the proposed dataset. The top row shows examples with an object within the gripper while the lower row corresponds to no object. Each example corresponds to a different batch.}
    \label{fig:synthetic_data}
\end{figure}

Fig.~\ref{fig:synthetic_data} shows six examples of rendered images of both classes. The gripper can be observed in several positions, with different distractor objects in the background. 

\section{GraspCheckNet}
The proposed approach for grasp verification is composed of a two-stage architecture that combines  object detection and image classification. 
Object detection is used to localize the robot's gripper within the image. This makes the architecture adaptable to different robotic platforms and object types, while the image classification model verifies whether the grasp was successful.
\subsection{Model Architecture}
GraspCheckNet consists of two primary components; a YOLO-based object detection model and a ResNet-based image classification model.
The object detection stage localizes the robot's gripper in the camera's field of view, while the classification model determines whether the gripper is holding an object.

The object detection model is a pretrained YOLO model fine-tuned on the HSR-GraspSynth dataset.
The image classification model is based on a ResNet~\cite{he_deep_2016} architecture, and operates on the detected region of interest containing the gripper produced by the detector.

The presence of an object is formulated as a binary classification task,  where a label of 0 indicates the presence of an object and a label of 1 signifies its absence. 

This labeling scheme aligns with our objective of detecting unsuccessful grasp attempts.

\begin{figure}[t]
    \centering
    \includegraphics[width=\linewidth]{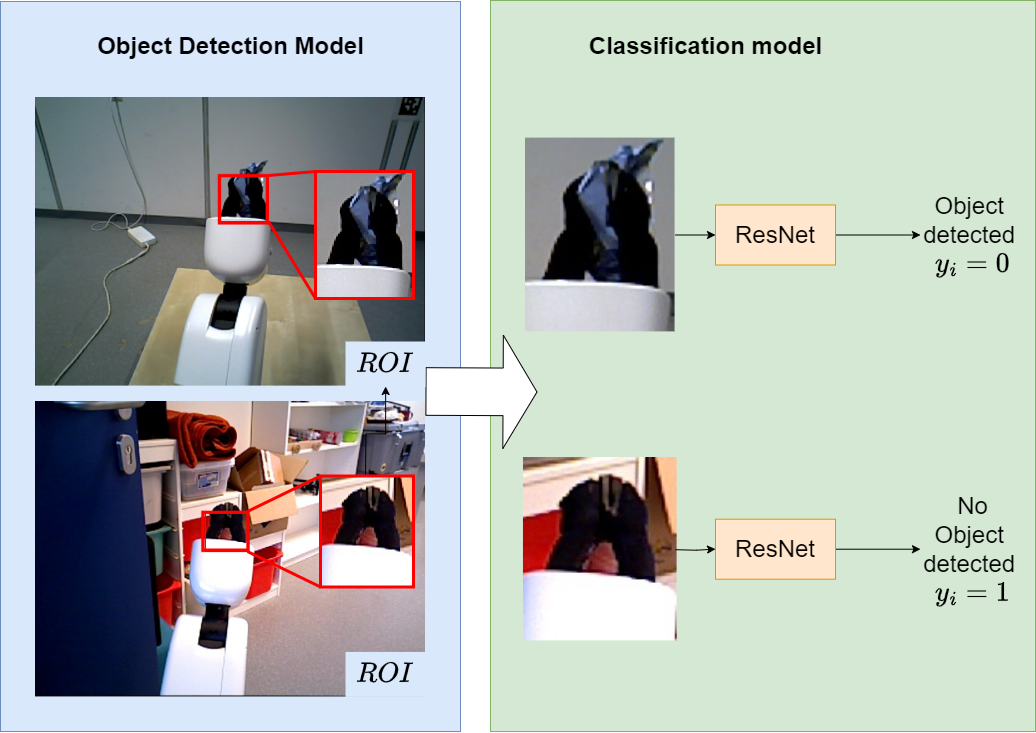}
    \caption{Illustration of the two-stage model architecture using object detection and image classification. The YOLO object detection model localizes the robot's gripper in the image, and the ResNet classification model uses the cropped image to determine whether there is an object in the gripper.}
    \label{fig:model_architecture}
\end{figure}

An overview of the model's architecture is shown in Fig.
~\ref{fig:model_architecture}.

The object detection step facilitates the classification task by eliminating irrelevant and unnecessary information from the image, focusing on the region of interest containing the gripper.
Alternatively, this stage could be replaced by a geometric-based approach if the robot's characteristics and kinematics are well-defined and accessible or by incorporating a predefined pose following the grasping procedure, where the robot positions the gripper directly in front of the camera. 
However, these alternative approaches require a higher degree of integration and interaction with the grasping pipeline, as they interrupt the grasping process and require additional time.

\section{EXPERIMENTS}
To evaluate the performance of the proposed grasp verification model and the accompanying dataset, we conduct experiments on both synthetic and real-world data. 
The primary objectives are to assess the model's effectiveness in accurately detecting the gripper and determining its state, as well as to evaluate the domain gap between the synthetic and real domain. 
Additionally, we compare it with an LLM-based Visual Question Answering approach as a few-shot alternative.

\subsection{Data Acquisition}
To assess the model's performance in real-world conditions, a smaller evaluation dataset of real-world images is created using the robot's onboard RGB camera.
Data collection is conducted in a room with furniture and domestic objects using Toyota's Human Support Robot (HSR).

The robot is placed in various environments with its arm extended, and its head-mounted camera oriented towards the gripper. 
Images are captured under different conditions, including scenarios where the gripper is empty and fully closed and others where it contains objects.  
A total of 518 real images, which we refer to as examples, are collected  distributed as follows:
\begin{itemize}
    \item 158 examples where the gripper is empty. 
    \item 150 examples where the gripper holds 16 different rigid objects. A comprehensive set of YCB-V objects and other household items found within the label are used.
    \item 210 examples where the gripper holds 23 different deformable objects. Various household items such as clothes, papers, chip bags and tissues are used.
\end{itemize}

Each object is captured between 5 and 10 times. 
The robot is placed in various locations.
The head is gradually rotated between consecutive captures of the same object to change the field of view and background.

\subsection{Object Detection Model}
We employ a fine-tuned YOLO11-l object detection model. 
The model is fine-tuned using Ultralytics' pretrained YOLO11-l~\cite{jocher_ultralytics_2023} on the proposed synthetic HSR-GraspSynth dataset.
To enhance the training process and mitigate the Sim-to-Real gap, various data augmentation techniques are applied during training.  
Used data augmentation techniques include  perspective and affine transforms,  and colour jitter, brightness and contrast changes and image compression.

The YOLO11-l model is fine-tuned for 100 epochs using Ultralytics' model  trainer with default parameters. 
The best model in terms of mean Average Precision (mAP) on the validation set is kept after the training process.

Due to the Sim-to-Real gap, a low confidence threshold is required during inference on real images, leading to a large number of candidate detections distributed across different clusters in the image.
To mitigate this issue, the confidence threshold is gradually reduced until detections appear.

\begin{figure}[t]
    \centering
    \begin{subfigure}{0.22\textwidth}
        \centering
        \includegraphics[width=\linewidth]{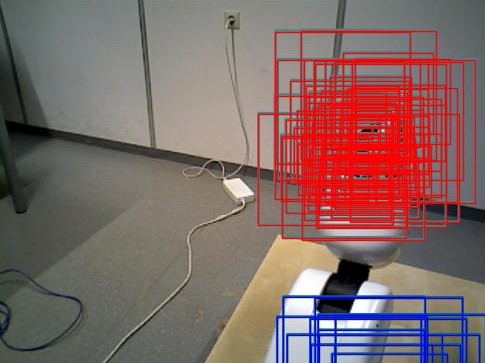}
    \end{subfigure}
    \hfill
    \begin{subfigure}{0.22\textwidth}
        \centering
        \includegraphics[width=\linewidth]{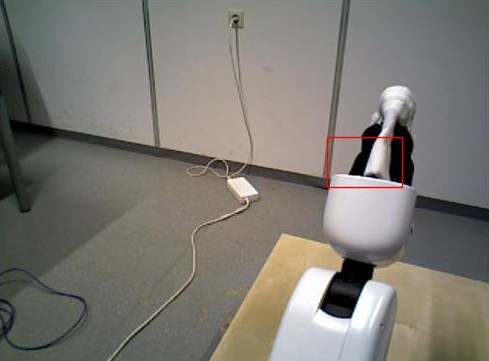}
    \end{subfigure}
    \caption{Illustration of the clustering procedure  applied to the detected bounding boxes. 
    DBSCAN is used to identify clusters and assign a cluster label to each bounding box (left). 
    Subsequently, the highest confidence bounding box from the cluster with the highest total confidence score is selected as the final detection (right).}
    \label{fig:cluster}
\end{figure}

When a low threshold value is used, a large number of detections localized around different clusters in the image can appear, 
leading to false positive detections.
To mitigate this, we implement a post-processing refinement step after detection.
Density-Based Spatial Clustering (DBSCAN) is used to identify clusters of detections within the image, as shown in Fig.~~\ref{fig:cluster}. 
DBSCAN is preferred to other clustering methods such as K-means because it does not require a predefined number of clusters.
The clusters are ranked according to the cumulative confidence scores of the bounding boxes they contain. 
The final detection is selected as the highest confidence bounding box within the  highest ranked cluster.

\subsection{Image Classification}
The image classification model is responsible for determining whether the gripper contains an object or is empty.
For this task, a pretrained ResNet-18 model is employed. 
Within the wider ResNet model family, the ResNet-18 was chosen for having fewer parameters than the bigger models of its family, making faster during training and inference.

The model's head is adapted for the task of binary classification. 
The original head is replaced by two fully connected layers with ReLU activation functions and dropout layers in-between~\cite{szeliski_computer_2022}. 

The model is trained using ground truth cropped synthetic images containing the robot's gripper.  
Data augmentation techniques are used to make the model robust to the synthetic to real domain transfer. 
The model is trained using an Nvidia A40 GPU in different stages. First, only the head is trained using a large dropout rate of 0.7 to 0.5 to make the model more robust. Afterwards, the learning and dropout rates are decreased while also unfreezing the backbone's last layer.

\subsection{Real-world evaluation}
To validate the model's effectiveness in real-world conditions, a qualitative evaluation is conducted. First, the object detection model (stage 1) is evaluated independently, followed by the evaluation of the classification module (stage 2) using the detections as input.

\subsubsection{Object Detection Model}
We first evaluate the ability of the detection model to localize the gripper within the image. 
Intersection over Union thresholds are not used for the assessment. 
Instead, the detections are qualitatively assessed based on whether the  bounding boxes sufficiently localize and encompass the robot's gripper. 
The fine-tuned YOLO model is used to obtain the gripper's bounding box for each of the 518 test images. 
Each detection is  manually reviewed and considered correct if it contains, at least partially, both gripper fingers and the majority of the bounding box area corresponds to the gripper and object, with limited inclusion of background regions.
Detections that mostly contain the background or fail to include both gripper fingers are labeled as incorrect.

Table~\ref{tab:real_object_detection_results} shows the results of this evaluation. We observe that the model is able to properly locate the gripper within the image in 98\% of \textit{no\_object} examples, 94.67\% of examples where there is a rigid object within the gripper and 96.67\% of  examples when there is a deformable object in the gripper.

\begin{figure}[t]
   \centering
    \begin{subfigure}{0.15\textwidth}
        \centering
        \includegraphics[width=\linewidth]{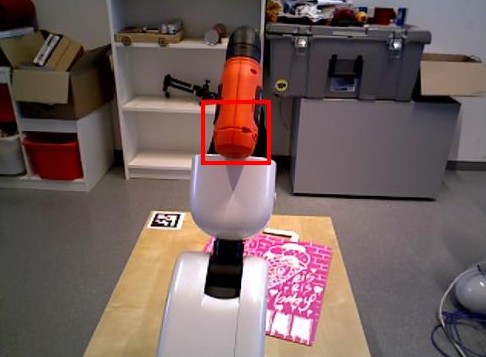}
    \end{subfigure}
    \hfill
    \begin{subfigure}{0.15\textwidth}
        \centering
        \includegraphics[width=\linewidth]{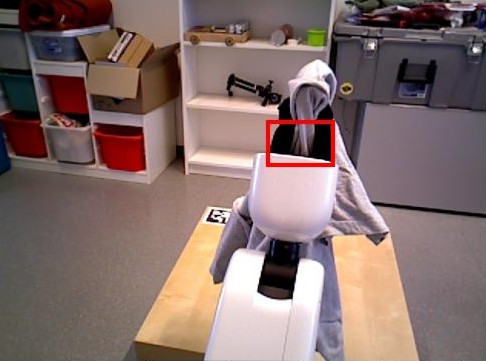}
    \end{subfigure}
    \hfill
    \begin{subfigure}{0.15\textwidth}
        \centering
        \includegraphics[width=\linewidth]{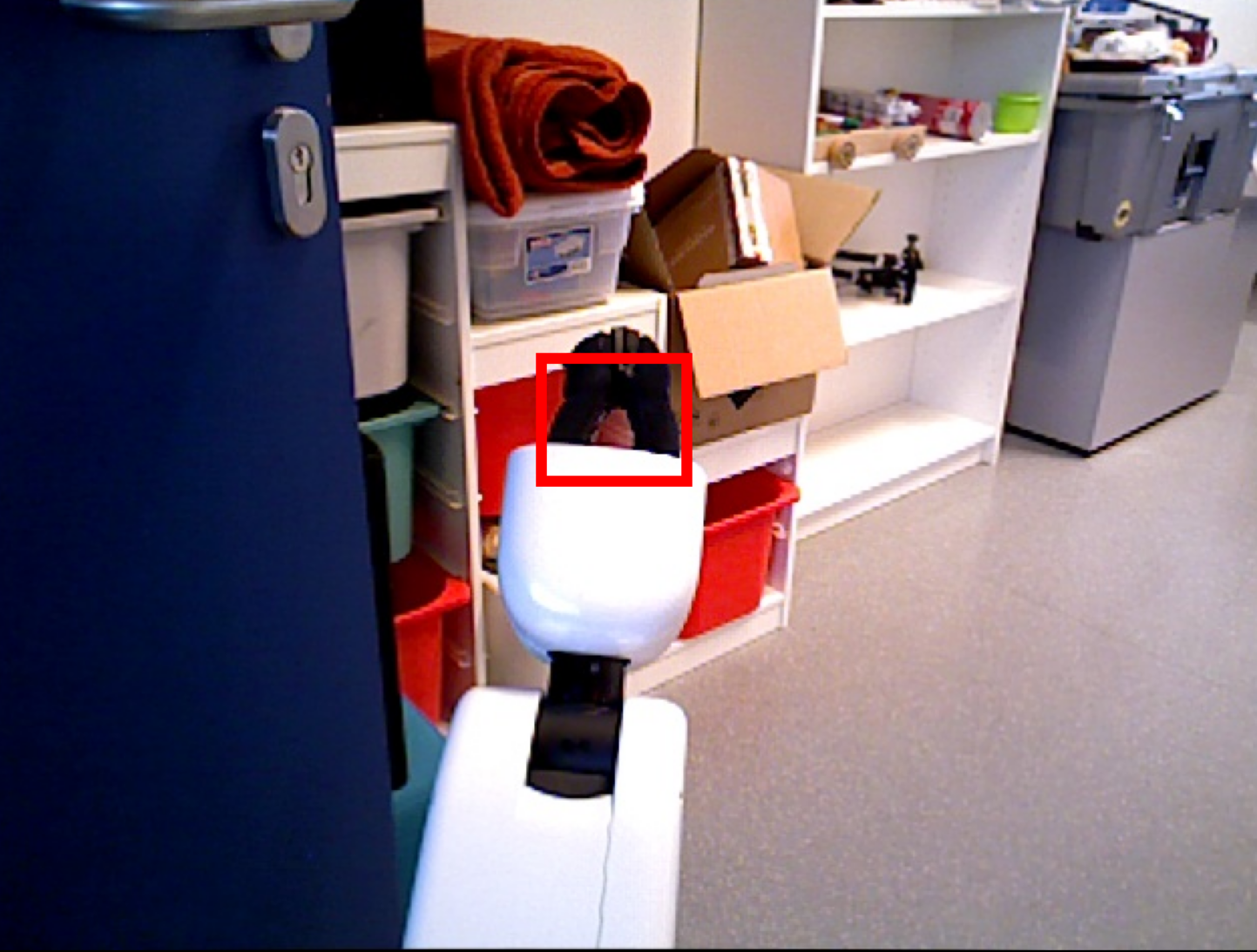}
    \end{subfigure}

    \vspace{0.5cm} 

    \begin{subfigure}{0.15\textwidth}
        \centering
        \includegraphics[width=\linewidth]{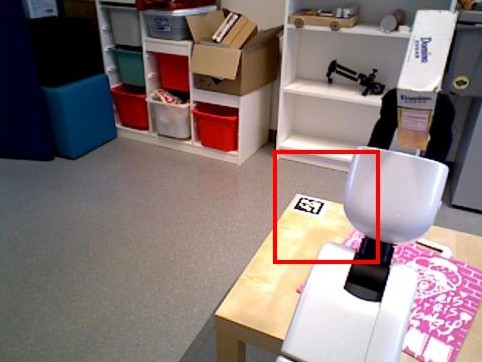}
    \end{subfigure}
    \hfill
    \begin{subfigure}{0.15\textwidth}
        \centering
        \includegraphics[width=\linewidth]{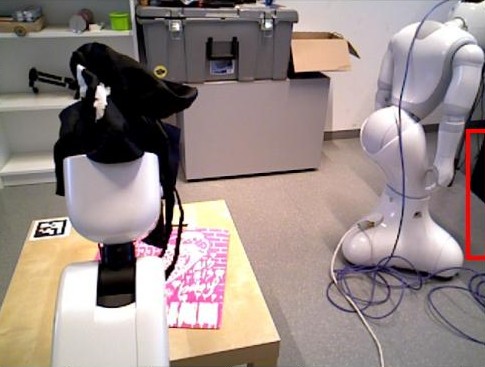}
    \end{subfigure}
    \hfill
    \begin{subfigure}{0.15\textwidth}
        \centering
        \includegraphics[width=\linewidth]{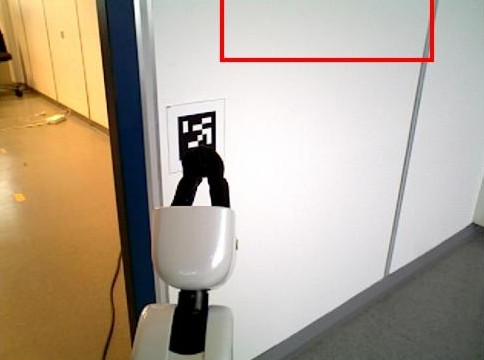}
    \end{subfigure}
    \caption{Sample detections using the object detection model. The red rectangle shows the detected bounding box. Top and lower rows show correct and incorrect detections respectively.}
    \label{fig:example_detections}
\end{figure}

Nevertheless, as shown in Fig.~\ref{fig:example_detections} the evaluation is limited by not taking into account the IoU. 
During the experiments it was observed that predicted bounding boxes tend to properly contain the gripper in terms of width, but often do not fully encompass it on the vertical axis.
During inference we mitigate it by padding the detected bounding box.

\begin{table}[b]
\caption{Evaluation of the object detection model on the real-world dataset. Num. detected refers to the number of examples where the predicted bounding box is qualitatively correct. Percentage of objects correct refers to individual objects that have been detected correctly in all the examples they apear in the dataset.}
\begin{center}
\begin{tabular}{|c|c|c|c|c|}
\hline
\textbf{Category} & \textbf{\shortstack{Num.  Images}}& \textbf{\shortstack{Num. \\Detected }}& \textbf{\shortstack{\% Detected}}& \textbf{\shortstack{\% Objects \\ Correct}} \\
\hline
No Object& 158  & 155& 98.10& N/A \\
Rigid & 150 & 142& 94.67& 62.50\\
Deformable & 210  & 203& 96.67& 82.61\\
\hline
\end{tabular}
\label{tab:real_object_detection_results}
\end{center}
\end{table}

\subsubsection{Image Classification Model}

The second stage of the model is evaluated using the cropped images obtained from the object detection outputs. Detected bounding boxes are used to crop the image region containing the gripper, with additional margins added to compensate for possible errors.

In order to account for the gap between synthetic and real data, the probability threshold for assigning to label 1 is lowered to 0.15 from 0.5 during evaluation. 

Table \ref{tab:real_object_classification} shows the evaluation metrics of the model across the \textit{no\_object} and \textit{object} categories, and differentiating by rigid and deformable object. While table ~\ref{tab:model_metrics} contains the precision and recall values for the task of detecting \textit{no\_object} instances.

The model has an accuracy of 74.7\% in examples where the gripper is empty and on 82.9\% of deformable object instances, indicating that it is able to properly recognize non-rigid objects within the gripper. 

In terms of detecting that the gripper is empty, as a class of binary classification, it achieves a precision value of only 0.678 albeit with a higher recall value. 
This lower precision value is indicative of the presence of false positives during the detection. 
In our case, in which we are mostly interested in detecting failed grasps, the focus is on obtaining a higher recall value.

In terms of execution time, the average inference time is of 28ms on an A40 GPU, when not taking into account the initial mode loading time. 
Extrapolating this result to less powerful devices indicates the suitability of the model for low-latency applications.

\begin{table}[!ht]
\centering
\caption{Classification accuracy per category. \textbf{Bold} indicates maximum, \underline{underline} indicates minimum performance.}
\renewcommand{\arraystretch}{1.2}
\begin{tabular}{|c|c|c|c|}
\hline
\textbf{Category} & \textbf{\shortstack{GraspCheckNet \\ Accuracy (\%)}} & \textbf{\shortstack{GPT-4o \\ Accuracy (\%)}} &\textbf{\shortstack{Llama 3.2 11B \\ Accuracy (\%)}}\\
\hline
No Object   & 74.7  & \textbf{95.0} & \underline{48.7} \\
Rigid       & 86.7  & \textbf{95.3} & \underline{68.7} \\
Deformable  & \textbf{82.9}  & 78.1 & \underline{60.0} \\
\hline
\end{tabular}
\label{tab:real_object_classification}
\end{table}

\begin{table}[!ht]
\centering
\caption{Precision and Recall score per model. \textbf{Bold} indicates maximum, \underline{underline} indicates minimum performance. }
\renewcommand{\arraystretch}{1.2}
\begin{tabular}{|c|c|c|}
\hline
\textbf{Model} & \textbf{Precision} & \textbf{Recall} \\
\hline
GraspCheckNet & 0.678 & 0.749\\
GPT-4o        & \textbf{0.739}& \textbf{0.95}\\
Llama 3.2     & \underline{0.357}& \underline{0.513}\\
\hline
\end{tabular}
\label{tab:model_metrics}
\end{table}

\subsection{Visual Question Answering}
In order to establish a baseline to which compare our GraspCheckNet model, we evaluate the use of state-of-the-art LLMs for Visual Question Answering as a zero-shot method for image classification. 
Our goal is to leverage their state-of-the-art performance in visual reasoning tasks to evaluate our model's performance.

We follow a visual-question-answering approach in which an LLM is prompted with the task that it should do and how to reply to it.  
The same prompt is used for all instances and no concrete information about the object in the gripper was included even though it might be available in certain grasping pipelines.
We evaluate two LLMs to compare the effects of the model size and whether on-device models are able to successfully complete the task. We test GPT4-o~\cite{openai_gpt-4o_2024} and llama 3.2 Vision 11B~\cite{meta_llama_2024}.
GPT4-o is tested using OpenIA's API while Llama 3.2 Vision is used through UnSloth's implementation of the model~\cite{daniel_han_unsloth_2023}, which reduces its memory footprint.
We use both the state of the art GPT4-o, which is closed-source and has large memory requirements, and Llama 3.2 in its 11B parameters version. 
This latter model can be run in consumer devices with approximately 6GB of GPU or unified system memory, making it feasible to deploy in practical scenarios.

Table \ref{tab:real_object_classification} shows the results of evaluating the VQA models on the real-world evaluation dataset.
Llama was not able to successfully perform the VQA task, achieving a recall of only 0.513. When asked concrete questions about the images, the model often produces hallucinations or does not correctly understand the scene. This indicates that further advancements in vision LLMs or the use of larger models is required.

On the other hand, GPT4-o is able to correctly detect most instances of the gripper being empty, with a recall of 0.95. 
However, it shows a relatively large amount of false positives in deformable objects, recognizing the gripper as empty.
This does not happen uniformly across all objects.
It is not able to properly detect some clothing items such as a black glove, a kitchen drape, a t-shirt and a hat, which account for 33 out of the 46 wrong classifications of deformable objects. 
These objects present uniform textures, without defining features, and when grasped by the gripper they do not hold a recognizable shape.
This might indicate that the vision model focus on the detection of an object and not in identifying whether there is anything in the gripper, making it susceptible to false positives when there are difficult to recognize objects.
It presents a higher amount of false positives, deformable object instances classified as empty, than our model while it has a lower amount of false negatives, instances of empty gripper classified as not empty.

The use of vision language models, even when using smaller models such as Llama 3.2 11B, requires expensive compute and requires more execution time than our proposed model. 
Open AI's GPT4-o required on average 2.27 seconds per image, albeit with a high standard deviation of 1.53 seconds. 
Due to the use of a remotely hosted API, the model's latency can often not be stable and a stable internet connection is required. 
The higher latency and requirement for internet connection makes this method less reliable for real-world applications.
In terms of cost, each instance costs approximately 0.001€ to execute. 
While this cost is relatively low, it can quickly scale up if a large amount of classifications is required.

When compared to our model, GraspCheckNet offers lower recall but can be run on-device with a lower inference time, making it feasible for low-latency applications and integration within grasping pipelines. Our model achieves comparable performance in detecting the presence of deformable objects but is less accurate to detect the gripper being empty.

\section{CONCLUSION AND FUTURE WORK}
This paper presents GraspCheckNet, a vision-based approach for grasp verification using head-mounted cameras, with a particular emphasis on deformable object manipulation. 
Our two-stage architecture uses object detection and image classification to verify successful grasps, addressing the challenges posed by non-rigid and deformable objects. We introduce HSR-GraspSynth, a synthetic dataset for training grasp verification models and help address the limitations of real-world data acquisition and reduce the Sim2Real gap.
Experimental results demonstrate that the proposed approach properly detects the presence of an object within the robot's gripper, particularly for deformable objects. 
Our approach maintains consistence performance while offering significant advantages in terms of inference and the ability to run on-device without requiring external APIs.

Future work should focus on the integration within grasping pipelines, exploring how real-time verification can be of use and how a better integration with the pipeline can be used to increase the model's accuracy. Furthermore, domain-adaptation techniques, both supervised and unsupervised could be explored to mitigate the Sim2Real gap.

%

\section*{ACKNOWLEDGMENT}
This research is supported by the EU program EC Horizon 2020 for Research and Innovation under grant agreement No. 101017089, project TraceBot, and the Austrian Science Fund (FWF), under project No. I 6114, iChores.

{
\small
\bibliographystyle{IEEEtranS}
\bibliography{references_zot}

\begin{thebibliography}{10}
\providecommand{\url}[1]{#1}
\csname url@rmstyle\endcsname
\providecommand{\newblock}{\relax}
\providecommand{\bibinfo}[2]{#2}
\providecommand\BIBentrySTDinterwordspacing{\spaceskip=0pt\relax}
\providecommand\BIBentryALTinterwordstretchfactor{4}
\providecommand\BIBentryALTinterwordspacing{\spaceskip=\fontdimen2\font plus
\BIBentryALTinterwordstretchfactor\fontdimen3\font minus \fontdimen4\font\relax}
\providecommand\BIBforeignlanguage[2]{{%
\expandafter\ifx\csname l@#1\endcsname\relax
\typeout{** WARNING: IEEEtran.bst: No hyphenation pattern has been}%
\typeout{** loaded for the language `#1'. Using the pattern for}%
\typeout{** the default language instead.}%
\else
\language=\csname l@#1\endcsname
\fi
#2}}

\bibitem{bicchi_robotic_2000}
A.~Bicchi and V.~Kumar, ``Robotic grasping and contact: a review,'' in \emph{Proceedings 2000 {ICRA}. {Millennium} {Conference}. {IEEE} {International} {Conference} on {Robotics} and {Automation}. {Symposia} {Proceedings} ({Cat}. {No}.{00CH37065})}, vol.~1, 2000, pp. 348--353 vol.1.

\bibitem{chang_shapenet_2015}
A.~X. Chang, T.~Funkhouser, L.~Guibas, P.~Hanrahan, Q.~Huang, Z.~Li, S.~Savarese, M.~Savva, S.~Song, H.~Su, J.~Xiao, L.~Yi, and F.~Yu, ``{ShapeNet}: {An} {Information}-{Rich} {3D} {Model} {Repository},'' Dec. 2015.

\bibitem{daniel_han_unsloth_2023}
\BIBentryALTinterwordspacing
M.~H. Daniel~Han and U.~team, ``Unsloth,'' 2023. [Online]. Available: \url{http://github.com/unslothai/unsloth}
\BIBentrySTDinterwordspacing

\bibitem{denninger_blenderproc_2019}
M.~Denninger, M.~Sundermeyer, D.~Winkelbauer, Y.~Zidan, D.~Olefir, M.~Elbadrawy, A.~Lodhi, and H.~Katam, ``{BlenderProc},'' Oct. 2019.

\bibitem{he_deep_2016}
K.~He, X.~Zhang, S.~Ren, and J.~Sun, ``Deep {Residual} {Learning} for {Image} {Recognition},'' in \emph{2016 {IEEE} {Conference} on {Computer} {Vision} and {Pattern} {Recognition} ({CVPR})}, June 2016, pp. 770--778.

\bibitem{howard_mobilenets_2017}
A.~G. Howard, M.~Zhu, B.~Chen, D.~Kalenichenko, W.~Wang, T.~Weyand, M.~Andreetto, and H.~Adam, ``{MobileNets}: {Efficient} {Convolutional} {Neural} {Networks} for {Mobile} {Vision} {Applications},'' Apr. 2017.

\bibitem{jocher_ultralytics_2023}
\BIBentryALTinterwordspacing
G.~Jocher, J.~Qiu, and A.~Chaurasia, ``Ultralytics {YOLO},'' Jan. 2023. [Online]. Available: \url{https://ultralytics.com}
\BIBentrySTDinterwordspacing

\bibitem{kulkarni_low-cost_2019}
P.~Kulkarni, S.~Schneider, and P.~G. Ploeger, ``Low-{Cost} {Sensor} {Integration} for {Robust} {Grasping} with {Flexible} {Robotic} {Fingers},'' in \emph{Advances and {Trends} in {Artificial} {Intelligence}. {From} {Theory} to {Practice}}, F.~Wotawa, G.~Friedrich, I.~Pill, R.~Koitz-Hristov, and M.~Ali, Eds.\hskip 1em plus 0.5em minus 0.4em\relax Cham: Springer International Publishing, 2019, pp. 666--673.

\bibitem{lin_explore_2023}
S.~Lin, K.~Wang, X.~Zeng, and R.~Zhao, ``Explore the {Power} of {Synthetic} {Data} on {Few}-shot {Object} {Detection},'' in \emph{2023 {IEEE}/{CVF} {Conference} on {Computer} {Vision} and {Pattern} {Recognition} {Workshops} ({CVPRW})}, 2023, pp. 638--647.

\bibitem{meta_llama_2024}
{Meta}, ``The {Llama} 3 {Herd} of {Models},'' 2024.

\bibitem{miller_graspit_2004}
A.~Miller and P.~Allen, ``Graspit! {A} versatile simulator for robotic grasping,'' \emph{IEEE Robotics \& Automation Magazine}, vol.~11, no.~4, pp. 110--122, Dec. 2004.

\bibitem{nair_combining_2017}
A.~Nair, D.~Chen, P.~Agrawal, P.~Isola, P.~Abbeel, J.~Malik, and S.~Levine, ``Combining self-supervised learning and imitation for vision-based rope manipulation,'' in \emph{2017 {IEEE} {International} {Conference} on {Robotics} and {Automation} ({ICRA})}, 2017, pp. 2146--2153.

\bibitem{nair_performance_2020}
D.~Nair, A.~Pakdaman, and P.~G. Plöger, ``Performance {Evaluation} of {Low}-{Cost} {Machine} {Vision} {Cameras} for {Image}-{Based} {Grasp} {Verification},'' Mar. 2020.

\bibitem{newbury_deep_2023}
R.~Newbury, M.~Gu, L.~Chumbley, A.~Mousavian, C.~Eppner, J.~Leitner, J.~Bohg, A.~Morales, T.~Asfour, D.~Kragic, D.~Fox, and A.~Cosgun, ``Deep {Learning} {Approaches} to {Grasp} {Synthesis}: {A} {Review},'' \emph{IEEE Transactions on Robotics}, vol.~39, no.~5, pp. 3994--4015, Oct. 2023.

\bibitem{openai_gpt-4o_2024}
\BIBentryALTinterwordspacing
{OpenAI}, ``{GPT}-4o {System} {Card},'' 2024, \_eprint: 2410.21276. [Online]. Available: \url{https://arxiv.org/abs/2410.21276}
\BIBentrySTDinterwordspacing

\bibitem{redmon_you_2016}
J.~Redmon, S.~Divvala, R.~Girshick, and A.~Farhadi, ``You {Only} {Look} {Once}: {Unified}, {Real}-{Time} {Object} {Detection},'' in \emph{2016 {IEEE} {Conference} on {Computer} {Vision} and {Pattern} {Recognition} ({CVPR})}, June 2016, pp. 779--788.

\bibitem{romano_human-inspired_2011}
J.~M. Romano, K.~Hsiao, G.~Niemeyer, S.~Chitta, and K.~J. Kuchenbecker, ``Human-{Inspired} {Robotic} {Grasp} {Control} {With} {Tactile} {Sensing},'' \emph{IEEE Transactions on Robotics}, vol.~27, no.~6, pp. 1067--1079, Dec. 2011.

\bibitem{sanchez_robotic_2018}
J.~Sanchez, J.-A. Corrales, B.-C. Bouzgarrou, and Y.~Mezouar, ``\BIBforeignlanguage{en}{Robotic manipulation and sensing of deformable objects in domestic and industrial applications: a survey},'' \emph{\BIBforeignlanguage{en}{The International Journal of Robotics Research}}, June 2018.

\bibitem{sun_accurate_2015}
L.~Sun, G.~Aragon-Camarasa, S.~Rogers, and J.~P. Siebert, ``Accurate garment surface analysis using an active stereo robot head with application to dual-arm flattening,'' May 2015, pp. 185--192.

\bibitem{szeliski_computer_2022}
R.~Szeliski, \emph{Computer vision: algorithms and applications}.\hskip 1em plus 0.5em minus 0.4em\relax Springer Nature, 2022.

\bibitem{tobin_domain_2017}
J.~Tobin, R.~Fong, A.~Ray, J.~Schneider, W.~Zaremba, and P.~Abbeel, ``Domain {Randomization} for {Transferring} {Deep} {Neural} {Networks} from {Simulation} to the {Real} {World},'' Mar. 2017.

\bibitem{tremblay_falling_2018}
J.~Tremblay, T.~To, and S.~Birchfield, ``Falling {Things}: {A} {Synthetic} {Dataset} for {3D} {Object} {Detection} and {Pose} {Estimation},'' July 2018.

\bibitem{weibel_measuring_2021}
J.-B. Weibel, R.~Rohrböck, and M.~Vincze, ``Measuring the {Sim2Real} {Gap} in {3D} {Object} {Classification} for {Different} {3D} {Data} {Representation},'' in \emph{Computer {Vision} {Systems}}, M.~Vincze, T.~Patten, H.~I. Christensen, L.~Nalpantidis, and M.~Liu, Eds.\hskip 1em plus 0.5em minus 0.4em\relax Springer International Publishing, 2021, pp. 107--116.

\bibitem{xiang_posecnn_2018}
Y.~Xiang, T.~Schmidt, V.~Narayanan, and D.~Fox, ``\BIBforeignlanguage{en}{{PoseCNN}: {A} {Convolutional} {Neural} {Network} for {6D} {Object} {Pose} {Estimation} in {Cluttered} {Scenes}},'' in \emph{\BIBforeignlanguage{en}{Robotics: {Science} and {Systems} {XIV}}}.\hskip 1em plus 0.5em minus 0.4em\relax Robotics: Science and Systems Foundation, June 2018.

\bibitem{yin_modeling_2021}
H.~Yin, A.~Varava, and D.~Kragic, ``Modeling, learning, perception, and control methods for deformable object manipulation,'' \emph{Science Robotics}, vol.~6, no.~54, 2021.

\bibitem{zhu_challenges_2022}
J.~Zhu, A.~Cherubini, C.~Dune, D.~Navarro-Alarcon, F.~Alambeigi, D.~Berenson, F.~Ficuciello, K.~Harada, J.~Kober, X.~Li, J.~Pan, W.~Yuan, and M.~Gienger, ``Challenges and {Outlook} in {Robotic} {Manipulation} of {Deformable} {Objects},'' \emph{IEEE Robotics \& Automation Magazine}, vol.~29, no.~3, pp. 67--77, Sept. 2022.

\end{thebibliography}
}

\end{document}